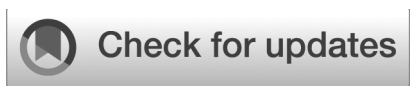

*CORRESPONDENCE
Kimberley S. Samkoe
✉ Kimberley.S.Samkoe@dartmouth.edu

# Fluorescence molecular optomic signatures improve identification of tumors in head and neck specimens

Yao Chen[1], Samuel S. Streeter[1,2,3], Brady Hunt[1], Hira S. Sardar[1], Jason R. Gunn[1], Laura J. Tafe[3,4], Joseph A. Paydarfar[3,5,6], Brian W. Pogue[1,3,7], Keith D. Paulsen[1,3] and Kimberley S. Samkoe[1,3,5]*

[1]Thayer School of Engineering, Dartmouth College, Hanover, NH, United States, [2]Department of Orthopaedics, Dartmouth Health, Lebanon, NH, United States, [3]Geisel School of Medicine, Dartmouth College, Hanover, NH, United States, [4]Department of Pathology, Dartmouth Health, Lebanon, NH, United States, [5]Department of Surgery, Dartmouth Health, Lebanon, NH, United States, [6]Department of Otolaryngology, Dartmouth Health, Lebanon, NH, United States, [7]Department of Medical Physics, University of Wisconsin-Madison, Madison, WI, United States

**Background:** Fluorescence molecular imaging using ABY-029, an epidermal growth factor receptor (EGFR)-targeted, synthetic Affibody peptide labeled with a near-infrared fluorophore, is under investigation for surgical guidance during head and neck squamous cell carcinoma (HNSCC) resection. However, tumor-to-normal tissue contrast is confounded by intrinsic physiological limitations of heterogeneous EGFR expression and non-specific agent uptake.
**Objective:** In this preliminary study, radiomic analysis was applied to optical ABY-029 fluorescence image data for HNSCC tissue classification through an approach termed "optomics." Optomics was employed to improve tumor identification by leveraging textural pattern differences in EGFR expression conveyed by fluorescence. The study objective was to compare the performance of conventional fluorescence intensity thresholding and optomics for binary classification of malignant vs. non-malignant HNSCC tissues.
**Materials and Methods:** Fluorescence image data collected through a Phase 0 clinical trial of ABY-029 involved a total of 20,073 sub-image patches (size of $1.8 \times 1.8$ mm$^2$) extracted from 24 bread-loafed slices of HNSCC surgical resections originating from 12 patients who were stratified into three dose groups (30, 90, and 171 nanomoles). Each dose group was randomly partitioned on the specimen-level 75%/25% into training/testing sets, then all training and testing sets were aggregated. A total of 1,472 standardized radiomic features were extracted from each patch and evaluated by minimum redundancy maximum relevance feature selection, and 25 top-ranked features were used to train a support vector machine (SVM) classifier. Predictive performance of the SVM classifier was compared to fluorescence intensity thresholding for classifying testing set image patches with histologically confirmed malignancy status.
**Results:** Optomics provided consistent improvement in prediction accuracy and false positive rate (FPR) and similar false negative rate (FNR) on all testing set slices, irrespective of dose, compared to fluorescence intensity thresholding (mean accuracies of 89% vs. 81%, $P = 0.0072$; mean FPRs of 12% vs. 21%, $P = 0.0035$; and mean FNRs of 13% vs. 17%, $P = 0.35$).
**Conclusions:** Optomics outperformed conventional fluorescence intensity thresholding for tumor identification using sub-image patches as the unit of analysis. Optomics mitigate diagnostic uncertainties introduced through physiological variability, imaging agent dose, and inter-specimen biases of fluorescence molecular imaging by probing textural image information. This preliminary study provides a proof-of-concept that applying radiomics to fluorescence molecular imaging data offers a promising image analysis technique for cancer detection in fluorescence-guided surgery.

## 1. Introduction

Head and neck squamous cell carcinomas (HNSCCs) account for approximately 900,000 cases worldwide and over 400,000 deaths annually (1). Surgical resection remains the first-line treatment for HNSCC. Cancer-free surgical margins with preservation of normal tissues are strongly associated with lower locoregional recurrence and improved patient outcomes (2). Unfortunately, achieving complete resection while minimizing unnecessary tissue removal is challenging due to tumor infiltration within sensitive head and neck anatomy. This clinical problem would benefit from precise intraoperative tumor identification and clear tumor-to-normal tissue differentiation during surgeries.

Over the past several years, near-infrared (NIR) fluorescence molecular imaging has entered the surgical theatre and shown promise for tumor identification during fluorescence-guided surgery (FGS) (3, 4). Optical tissue contrast is provided by differential uptake of the fluorescent agent in the tumor relative to surrounding normal tissues. Additionally, molecularly targeted fluorescent agents bind preferentially to receptors overexpressed in tumors, further increasing the fluorescence signal in cancers. Epidermal growth factor receptor (EGFR), a transmembrane cell surface glycoprotein associated with cell proliferation (5), is commonly expressed at high levels in a variety of epithelial tumors, including HNSCCs (>90%) (6–8), providing a strong rationale for investigating imaging agents that target this receptor (9–11). Several EGFR-targeting agents have been studied in clinical trials and used in a variety of surgical applications (12–15).

In practice, EGFR-targeted fluorescence imaging achieves modest optical tumor-to-normal ratios (16), which are not always representative of actual differences in cellular expression of EGFR among tumor and normal tissues (14, 17–19) due to confounding effects involved in the entire fluorescence process from illumination to detection. For example, noise and signal loss are introduced by imaging instruments, fluorophore photochemical events, and tissue-photon interactions (20). Several groups have developed methods to overcome these optical limitations through instrument and illumination optimization (21, 22). Physiologically, variability results from non-specific agent uptake in normal tissues (i.e., non-specific binding in normal tissues with endogenously high EGFR expression like mucosa, salivary glands, and tonsils, and non-specific retention and perfusion of the agent caused by differences in vascular permeability) (23, 24) or from limited uptake in tumor tissues (i.e., not all HNSCCs overexpress EGFR due to heterogeneous gene expression, or complex, macroscopic structures reduce agent diffusion in regions of necrosis) (25). To address these physiological effects, explorations of new probes (19, 26, 27), dosing strategies (18, 28–30), and innovative methods for quantitative molecular imaging (31–35) have increased tumor contrast. Computational image analyses also address imaging and physiological impacts on tumor contrast (36).

ABY-029 is composed of an anti-EGFR Affibody molecule (Affibody AB, Solna, Sweden), a 58-amino acid synthetic peptide (Z03115), and IRDye800CW (LI-COR Biosciences, Inc., Lincoln, NE) (37). This study was motivated by observations of distinct spatial patterns in EGFR-targeted fluorescence of ABY-029 in tumor vs. normal tissues in HNSCC specimens, as illustrated in **Figure 1**. The effect was corroborated by inspection of EGFR immunohistochemistry (IHC)-stained tissues that were spatially co-registered to fluorescence images (35). In general, the distribution of EGFR expression in normal tissues appears homogeneous in intensity and spatially ordered, whereas it demonstrates heterogeneous staining intensity and spatial disorder in tumor tissues. Leveraging these distinct spatial distributions of EGFR expression may yield contrast-enhanced visualization of tumor in HNSCCs relative to using fluorescence intensity alone, which by itself, does not incorporate spatial relationships or patterns in local fluorescence.

Radiomics is the high-throughput extraction and characterization of quantitative features from medical image data (38–40). The approach generates first-order histogram statistics (i.e., intensity metrics) in addition to second- and higher-order pixel statistics (i.e., textures) (39). Radiomics also mines image features extracted for artificial intelligence-based decision support (41, 42) and has been used to detect disease and/or stratify tissues in conventional medical imaging data (e.g., magnetic resonance imaging, computed tomography) (43–45). Streeter et al. applied a radiomics framework to multi-wavelength optical spatial frequency domain imaging (SFDI) of breast cancer tissues (46). Through extension of the "-omics" concept to high-dimensional optical imaging data, the authors proposed the term "optomics" and developed a radiomics/optomics classification paradigm based on multimodal, micro-computed tomography and SFDI data that showed multimodal image feature sets improved detection of breast cancer samples compared to either modality alone. Radiomics is based on the premise that large numbers of quantitative images features can be mined and related to disease phenotypes as representations of their underlying genetic expression (43). In similar fashion, applying optomics to EGFR-targeted fluorescence images may reveal spatial patterns in EGFR expression that are associated with native tissue molecular signatures.

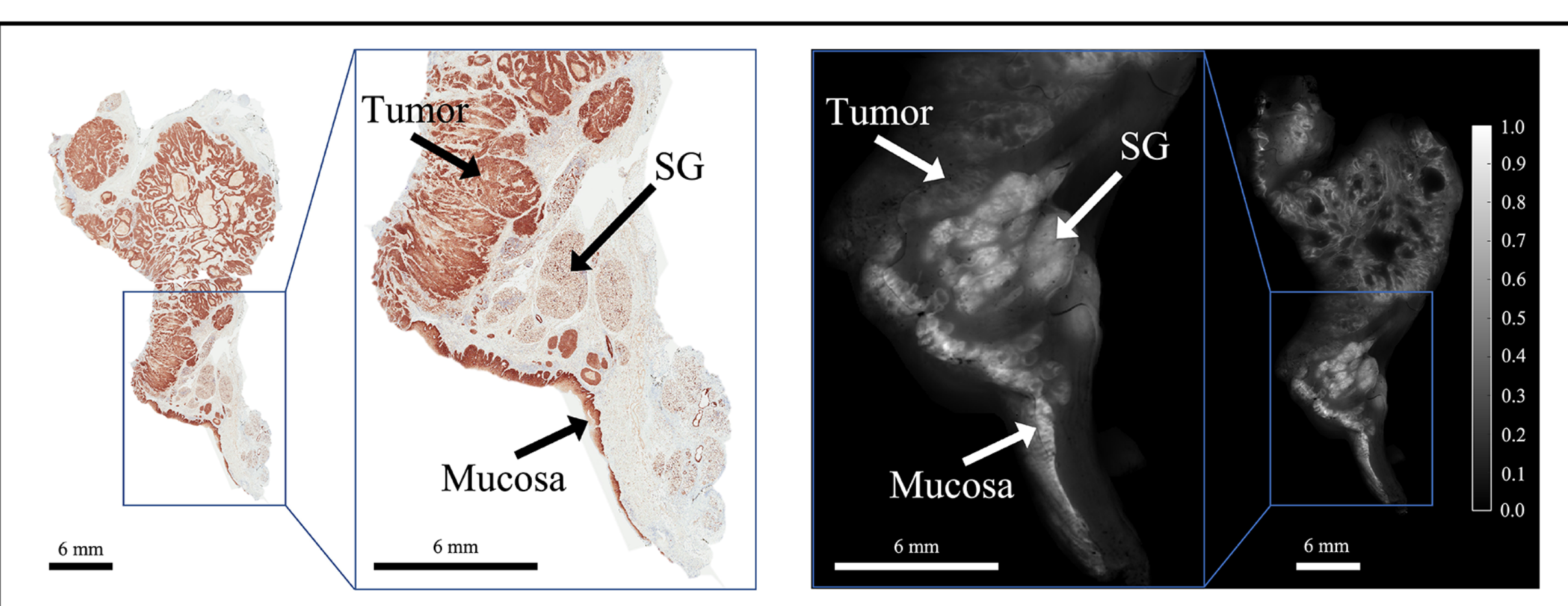

FIGURE 1
Representative bread-loafed slice of a head and neck cancer displaying normal and tumor-derived EGFR expression. EGFR IHC staining (*left*) and fluorescence (*right*) images of slice #16 from dose group 2. Arrows annotate EGFR-positive tissues including tumor, salivary gland (SG), and mucosa on the EGFR IHC staining and fluorescence images, respectively.

In this study, an optomics analysis framework was developed to classify HNSCCs and normal tissues using EGFR-targeted fluorescence images of human surgical specimens derived for presurgical administration of ABY-029. The work evaluated the predictive performance of conventional intensity-based thresholding and optomics for identifying malignant HNSCC tissues in resected surgical specimens.

# 2. Materials and methods

## 2.1. Clinical trial design

Data evaluated in this paper were acquired during a first-in-human, Phase 0, open-label, single-center, clinical trial of microdose evaluation of ABY-029 (eIND application 122681) in head and neck cancer (ClinicalTrials.gov identifier: NCT03282461). The primary goal of the trial was to determine if microdose, or near microdose, administration ABY-029 led to detectable fluorescence signals. The trial protocol was approved by the Dartmouth Health Institutional Review Board, and written informed consent was obtained from all patients. Additional information related to trial eligibility criteria and the imaging protocol can be found elsewhere (47).

Enrolled patients ($n = 12$) were assigned to one of three dose groups: Group 1 ($n = 3$) received 30 nanomoles/patient; Group 2 ($n = 3$) received 90 nanomoles/patient; and Group 3 ($n = 6$) received 171 nanomoles/patient. **Table 1** summarizes the clinical trial dose groups and patient characteristics.

## 2.2. Post-resection surgical specimen processing

The gross primary tumor specimens were sectioned into ~5-mm thick bread-loafed slices according to standard-of-care. Most slices (22 out of 24) in this study are a heterogeneous mix of tumor and normal tissues whereas the remaining two slices contained exclusively tumor. Each bread-loafed specimen was imaged using the 800 nm channel of the Odyssey CLx (LI-COR Biosciences, Inc., Lincoln, NE) with 42-µm pixel resolution. Formalin-fixed, paraffin-embedded blocks were created and sectioned (4 µm thickness). Hematoxylin and eosin (H&E) and EGFR IHC slides were obtained from consecutive 4-µm sections taken from the imaged tissue surface. The histology slides were examined by a board-certified pathologist (L.J.T.), who delineated tumor and non-malignant regions of interest (ROIs) on each slide. Slides were imaged with and without the ROIs at 20× magnification using an Aperio AT2 DX System (Leica Biosystems, Wetzlar, Germany). ROIs were co-registered with the fluorescence images by overlaying the greyscale fluorescence and H&E images and digitally annotating the former with respect to the latter using a custom MATLAB script (vR2021b, MathWorks, Natick, MA). **Figure 2** illustrates the clinical workflow for an enrolled study subject and includes representative images of surgical specimens at each stage of specimen processing.

TABLE 1 Dose groups and patient characteristics.

| | Patient index | Age | (M)ale/(F)emale | Cancer site | Number of bread-loafed slices |
|---|---|---|---|---|---|
| Group 1 (30 nanomoles/patient) | 1 | 45 | M | Right tonsil, base of tongue | 6 |
| | 2 | 59 | M | Oropharynx—soft palate | 2 |
| | 3[a] | 60 | M | Tongue | 0 |
| Group 2 (90 nanomoles/patient) | 4 | 55 | F | Buccal mucosa | 3 |
| | 5 | 56 | F | Tongue | 4 |
| | 6 | 50 | F | Retromolar trigone | 1 |
| Group 3 (171 nanomoles/patient) | 7 | 60 | M | Floor of mouth | 1 |
| | 8 | 59 | M | Tongue | 1 |
| | 9 | 82 | F | Buccal mucosa | 2 |
| | 10 | 45 | F | Retromolar trigone | 1 |
| | 11 | 48 | M | Oropharynx | 1 |
| | 12 | 86 | M | Tongue | 2 |

[a]Surgical specimens collected from the third patient did not have corresponding pathology images for later analysis.

TABLE 2 Sampled patch counts (size of 1.8 × 1.8 $mm^2$) in the training set by tissue malignancy.

| Tissue malignancy | Patch count | Patch counts from each specimen (mean ± 1 std) | Histopathologically confirmed ROI size for each specimen (mean ± 1 std, $mm^2$) |
|---|---|---|---|
| Malignant | 2,735 | 204 ± 167 | 62.9 ± 58.3 |
| Non-malignant | 11,928 | 567 ± 398 | 270.3 ± 173.6 |
| *Total* | 14,663 | – | – |

## 2.3. Design of image classification techniques

### 2.3.1. Image pre-processing

Each fluorescence image was background subtracted and normalized to a fluorescence calibration target placed in the field-of-view of every image. The process eliminated day-to-day instrument-specific variations (48, 49), which would confound comparisons of fluorescence data obtained during the trial. The background-subtracted and calibrated image data were then rescaled such that all ROI pixels fell in the intensity range of [0,1], which eliminated the inter-slice fluorescence bias caused by variations in imaging agent dose and heterogeneous patient characteristics. The image normalization step was also necessary for optomic feature extraction which required a consistent range of intensities across image samples. **Figure 3** shows representative examples of the histological and fluorescence image data analyzed, including three bread-loafed slices from each dose group, displayed both as the raw images using the same intensity scale and as standardized fluorescence images in which the intensities should be uniform.

### 2.3.2. Data partitioning strategy

Data were partitioned at the specimen-level (24 fluorescence bread-loafed specimens, one image per specimen, 8 specimens in each dose group) into training and testing sets. Slices were distributed randomly into training/testing sets with a 75%/25% split (6 and 2 slices in training and testing sets, respectively) for each dose group. After partitioning each dose cohort separately, all training and testing sets were aggregated. Thus, a total of 18 and 6 specimens comprised the training set and testing set, respectively. The same training and testing sets were used by each classification method (i.e., intensity thresholding and optomics). **Figure 4** summarizes the data partitioning and indexing of bread-loafed specimens in the training/testing sets along with the workflows of two image classification methods.

### 2.3.3. Fluorescence intensity thresholding method

The fluorescence intensity thresholding method was chosen as the baseline method to compare to optomics method. It used pixel intensity values alone as the discriminator to perform binary classification. The approach is a conventional, widely adopted method to perform binary classification of tissue types using fluorescence imagery, with the assumption that malignant tissues exhibit increased fluorescence intensity relative to non-diseased tissues (16, 50–54). Receiver operator characteristic (ROC) curves were created using image pixels as the unit of analysis. Pixels within the histologically confirmed ROIs of tumor that exhibited intensities above/below the threshold were classified as true positives/false negatives. Pixels contained within the normal tissue ROIs that exhibited intensities above/below the threshold were classified as false positives/true negatives. The threshold was swept through all possible intensities, and an optimum cutoff point (OCP) was selected based on the maximum pixel-level classification accuracy achieved on the training set. **Figure 4B** describes the intensity-based thresholding method performed in this study (blue boxes).

### 2.3.4. Optomics method

Optomic features were computed on the sub-image (or patch) level. Sub-image sampling was implemented prior to feature extraction to enlarge the dataset for model training and enabled patch-level classification. The sub-image sampling process involved randomly selecting center pixels in the histopathologically confirmed tissue ROIs and expanding to 1.8 mm × 1.8 mm (43 pixel by 43 pixel) patches from each center pixel. At least 90% patch area was required to be inside the ROIs to ensure sampled patches did not contain too many tissues outside the ROIs, leading to a sparse patch distribution near the edge region of each ROI.

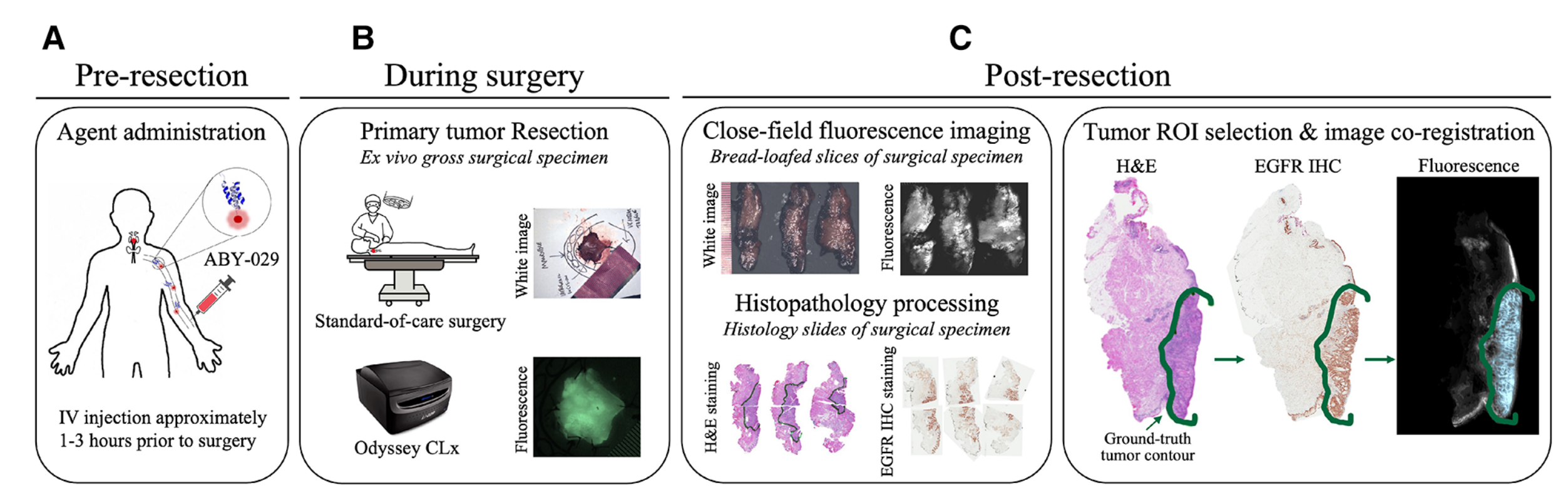

FIGURE 2
Clinical trial workflow. (A) ABY-029 was administered *via* 5 ml intravenous bolus injection 1–3 h prior to surgery. (B) Each patient underwent standard-of-care surgery, and gross surgical specimens were collected. (C) After surgery, specimens were sectioned in bread-loafed slices, formalin-fixed, and paraffin-embedded, then cut into consecutive slides for staining. Closed-field fluorescence imaging was performed on the bread-loafed slices. Histology slides from bread-loafed slices underwent standard histopathological examination, providing tissue ROI ground truth.

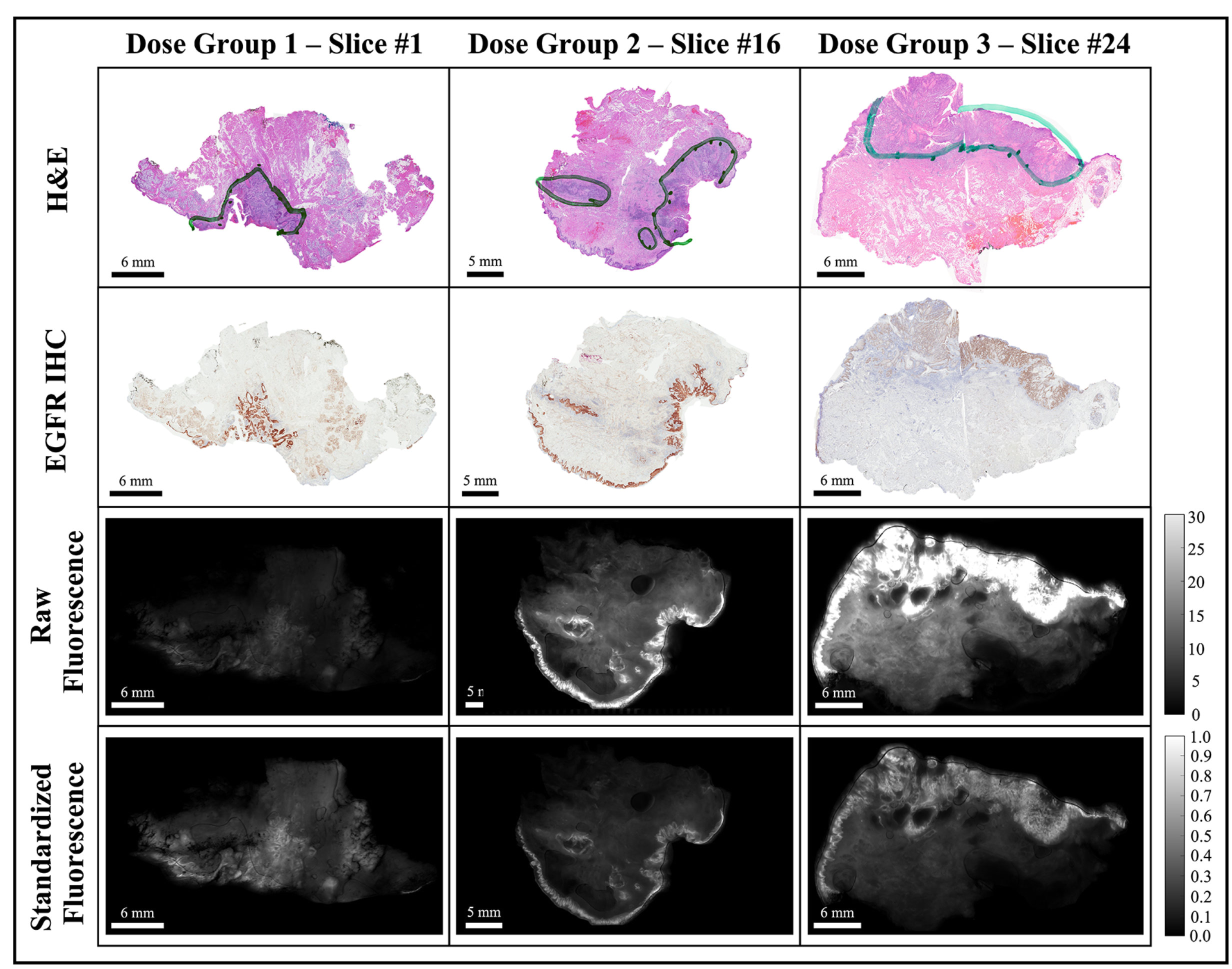

FIGURE 3
Representative bread-loaf slices from the three dose groups (30, 90, and 171 nanomoles). Green contours in the H&E images indicate ground truth tumor regions delineated by an expert pathologist (L.J.T.). The bottom two rows show comparisons of fluorescence images before (raw fluorescence) and after image pre-processing (standardized fluorescence).

Table 2 lists the number of sampled training set image patches by malignancy status. Meanwhile, testing set slices were sampled randomly over the entire tissue surface, without constraint to specific tissue ROIs.

After sub-image sampling, optomic features were extracted using the *PyRadiomics* package (v3.0.1) (55), which is open-source and Image Biomarker Standardization Initiative (IBSI)-compliant (56). A total of 1,472 *PyRadiomics* features were extracted from each sampled patch. Features included first-, second-, and higher-order pixel statistics (41): 18 first-order features described the distribution of intensities of individual pixels, and 74 second- and higher-order features were quantified, which captured the statistical interrelationships between sets of two or more pixels. In addition to extracting these 92 features from the original patch, fifteen image filters were applied and the same 92 features were extracted from each filtered patch. In summary, each wide-field fluorescence image of a bread-loafed tissue specimen was represented by a group of sub-images, and each sub-image was characterized by a total of 1,472 optomic features (92 original image patch features + 15 filters × 92 features/filter). Every feature across all patches was standardized to Z-scores prior to model training. Figure 4B illustrates the steps involved in the optomics method (orange boxes). Additional details related to the *PyRadiomics* features used are described in the Supplementary Material.

A supervised machine learning (ML) pipeline was developed using only training set image patches. Three hyperparameters in the ML pipeline were tuned *via* grid search: the ML classifier, feature selection algorithm, and number of top-ranked optomic features to include in model training. A total of seven ML classifiers and seven feature selection algorithms were evaluated based on their popularity, simplicity, and computational efficiency, as reported in the literature (57, 58). Classifiers included Random Forest (RF), *k*-Nearest Neighbors (*k*NN), Decision Tree (DT), Support Vector Machine (SVM), Boosting (BST), Bayesian (BY), and Discriminant Analysis (DA). Feature selection algorithms were minimum redundancy maximum relevance (MRMR), Fisher score ranking (FSCR), Chi-square ranking (CHSQ), Gini index ranking (GINI), mutual information maximization (MIM), Spearman ranking correlation coefficient (SRCC), and Pearson ranking correlation coefficient (PRCC). The number of top-ranked features ranged from 5 to 100 features in increments of 5 [i.e., $(5k \mid k \in \mathbb{Z}, 1 \leq k \leq 20)$]. Leave-One-Out Cross-Validation (LOOCV) on the specimen-level was performed on the training set (i.e., 18 folds) for hyperparameter tuning. LOOCV holds a group of patch feature data from a single specimen as the validation set for each split. LOOCV mean accuracy of each hyperparameter combination was obtained by averaging the validation specimen accuracies across all splits. The hyperparameter combination that generated the highest LOOCV mean accuracy was chosen as the final ML pipeline for model training and testing. Each combination of hyperparameters was evaluated through LOOCV of the training set (i.e., 7 classifiers $\times$ 7 feature selection methods $\times$ 20 options of feature numbers = 980 combinations). The optimal number of selected features was defined as the fewest number of features to be within 1.5 standard deviations of the maximum mean validation accuracy. The process of determining the best performing hyperparameter combination is summarized in Supplementary Figure S1. The final ML pipeline was evaluated on image patches extracted from the tissue ROIs in the 6 testing set specimens.

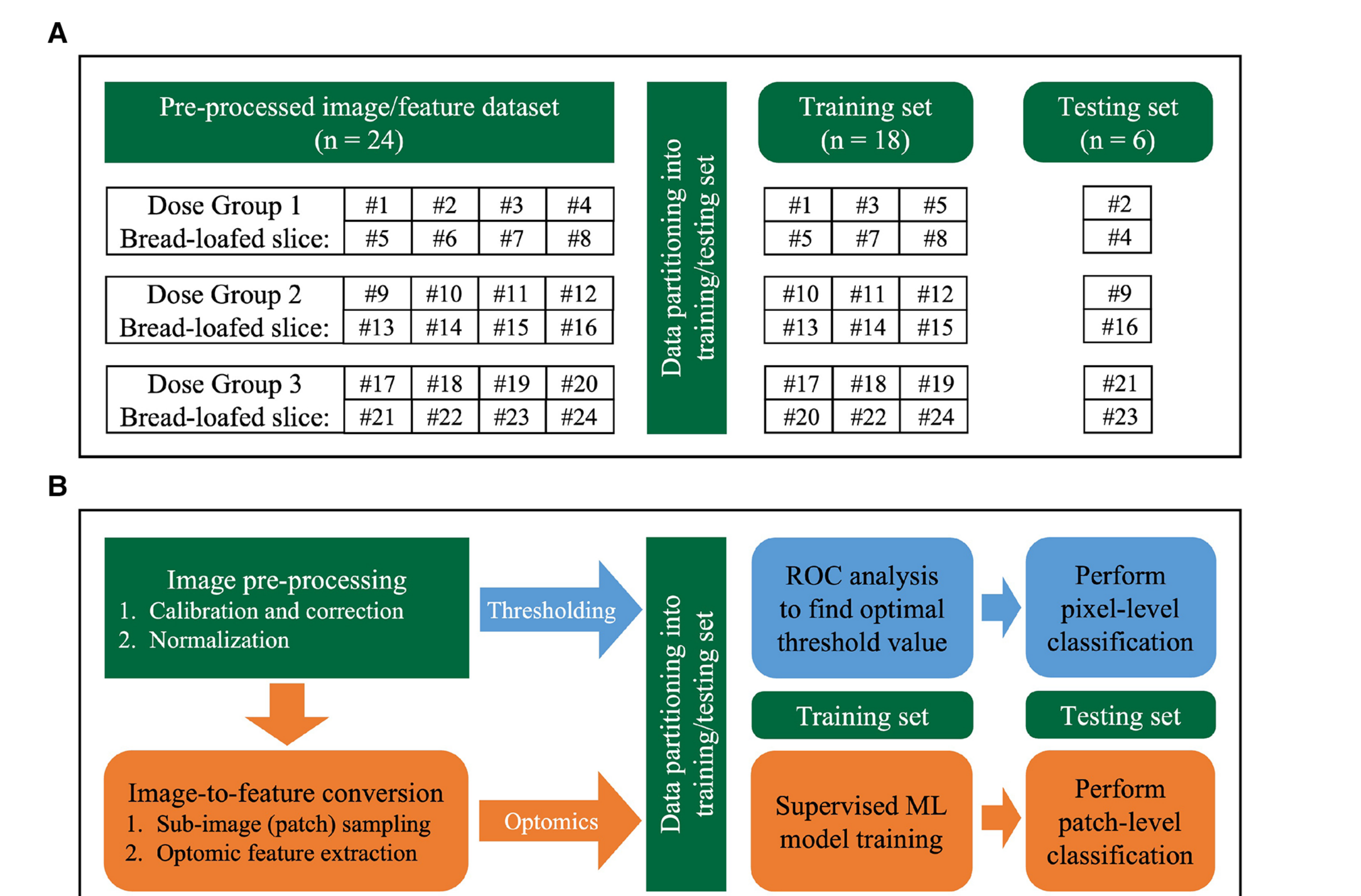

FIGURE 4
Image analysis workflow. (A) Data partitioning strategy. Bread-loafed slices were partitioned 75%/25% into aggregated training and testing sets. The same specimen-level partitioning strategy was used to assess the performance of both methods. (B) Workflow comparison of thresholding and optomics methods. Green modules indicate steps shared by the two methods. Blue and orange modules represent the steps unique to the thresholding and optomics approaches, respectively.

## 2.4. Statistical analysis

All statistical analysis was implemented in MATLAB (vR2021b, MathWorks, Natick, MA). A two-sided, paired *t*-test evaluated statistical differences between classification accuracy, false positive rate (FPR), and false negative rate (FNR) achieved by the two classification methods (i.e., thresholding vs. optomics) on the 6 testing set specimens. A *P*-value of 0.05 or less was considered statistically significant.

# 3. Results

## 3.1. ML pipeline hyperparameter optimization

Figure 5A depicts the LOOCV mean accuracies achieved by every combination of ML classifier type (rows) and feature selection algorithm (columns) when 25 top-ranked features were used. (Heatmaps for different numbers of selected features were similarly produced in this two-dimensional format, leading to a three-dimensional LOOCV mean accuracy representation over all combinations of the three hyperparameters; see Supplementary Figures S2A–2S.) The combination of SVM classifier and MRMR feature selection algorithm yielded the overall highest accuracy, as shown in Supplementary Figure S2T. Mean CV accuracy increased as the number of features increased, as displayed in Supplementary Figure S2U. The optimal hyperparameter combination was a SVM classifier, MRMR feature selection

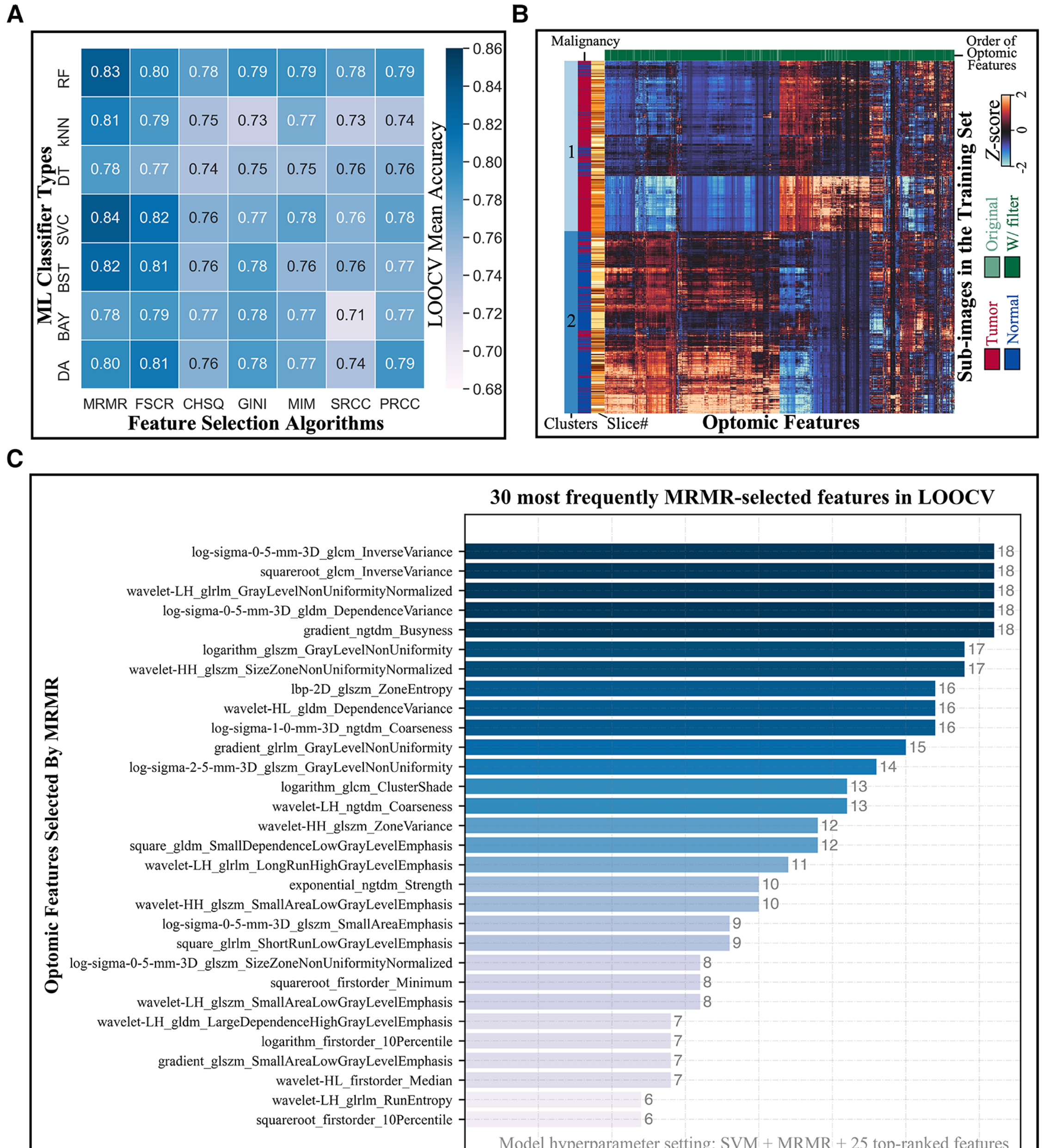

FIGURE 5
Model hyperparameter optimization through LOOCV and feature selection. (A) Heatmap depicts the LOOCV mean accuracy achieved by different combinations of classifiers (rows) and feature selection algorithms (columns) when 25 top-ranked features were used. The combination of SVM classifier and MRMR feature selection showed the highest validation accuracy. (B) Hierarchical clustering of training set sub-image patches (5,470 total; vertical axis) and optomic features (1,472 total; horizontal axis) reveals natural groupings of tissue types. (C) Most frequently selected features by MRMR with the optimal hyperparameters combination during LOOCV. Horizontal axis represents the number of times that each optomic feature was selected by MRMR in all 18 folds of LOOCV.

algorithm, and 25 top-ranked features, which provided a LOOCV mean accuracy of 84% ± 13%.

Figure 5B shows a hierarchically clustered heatmap of all optomics features extracted from training set patches (standardized to Z-scores). Two dominant clusters of sub-image samples are apparent: one containing mainly normal samples and one containing mainly tumor samples (84% normal in Cluster 1; 82% tumor in Cluster 2), which suggests that optomic feature sets extracted from fluorescence images have potential to classify tissue malignancy with high fidelity.

Figure 5C shows the distribution of tabulated optomic features selected by MRMR in each of 18 LOOCV folds. The most frequently selected features were all filter-based features, and all but four were second- or higher-order statistics, indicating that textural information was favored by MRMR feature selection for separating malignant from non-malignant samples.

## 3.2. Comparison of thresholding and optomics classification methods

For the fluorescence intensity thresholding method, the OCP value determined on the training set through ROC curve analysis was 0.155, as shown in Figure 6A. This OCP was used as the threshold to classify every ROI pixel in the 6 test set specimen images. Threshold classification results, based on co-registered, pathologist-annotated H&E images, were an accuracy of 79.7%, FPR of 17.5%, and FNR of 25.0%. Figure 6A shows typical thresholding results on a representative specimen in the test set.

Optomics classification performance on the same representative test set specimen appears in Figure 6B. Optomics achieved an accuracy of 93.5%, FPR of 8.8%, and FNR of 2.5%. (Visual comparisons of classification performance by the two methods for all other test set specimens are displayed in Supplementary Figure S4.)

Optomics provided consistent improvement on all 6 testing set slices in predication accuracy, FPR and similar FNR relative to thresholding (mean accuracies of 89% vs. 81%, mean FPRs of 12% vs. 21%, and mean FNRs of 13% vs. 17%,). A statistically significant difference in test set prediction accuracy and FPR was found ($P = 0.0072$ for accuracy and $P = 0.0035$ for FPR respectively), but no statistically significant difference was found for prediction FNR. Figure 7 compares prediction accuracies of thresholding and optomics classification methods on all six testing set specimens. Additionally, the interquartile range of thresholding prediction accuracies was larger than the optomics method (threshold method range of 73.5%–87.2% vs. optomics method range of 82.6%–93.5%).

## 3.3. Optomics method prediction probability map

To visualize optomics method prediction results on a pixel-level, a pseudocolor-coded heatmap (54) of model prediction probability was overlaid on each original greyscale fluorescence image. The continuous, pixel-level visualization was generated through biharmonic spline interpolation of the patch-level model probabilities assigned to the patch center pixel (using biharmonic spline interpolation*; v4, griddata*, implemented in MATLAB, vR2021b, MathWorks, Natick, MA). Figure 8A demonstrates the process of assigning the patch-level predicted probabilities to patch center pixels, and Figure 8B shows a pseudocolor visualization of the probability map overlaid on the greyscale fluorescence image.

To reduce bias introduced by assigning the patch-level predicted probability to the center pixel, independent ML models (with optimized hyperparameters) were generated from square patches with three different sizes (1.8-, 1.4-, and 0.9-mm side length) surrounding the same center pixel (Figure 8C, see Supplementary Figure S5 for ML model process for different patch sizes). The final center pixel probability was determined by averaging the predicted probabilities from the ML models with the three patch sizes. Figure 8D shows the interpolated probability map from the prediction outputs from the three patch sizes in comparison to Figure 8B generated from a single patch size (1.8 mm × 1.8 mm). The number of pixels with high tumor probability (visualized in yellow) but outside the ground truth tumor ROI decreased, while the number of those inside the ground truth tumor ROI increased when multiple patch sizes were included in the analysis. Optomics method prediction probability maps of the other five testing slices are shown in Supplementary Figure S6.

# 4. Discussion

In this study, classification performance of conventional fluorescence intensity thresholding was directly compared to an optomics approach for malignant HNSCC tissue detection. Optomics leverage high-throughput feature extraction to derive textural image information which reflects pixel relationships to structural tissue characteristics that have been shown to provide high prognostic power in other radiomic settings (59–62). The thresholding method is the most direct and commonly used binary classification method to distinguish tumor and normal tissues by the magnitude of fluorescence intensity, so we choose it as the baseline method to compare to optomics. The thresholding method provides limited tumor-to-normal tissue differentiation, since it ignores the fluorescence spatial patterns that may be related to tissue types or disease states.

The fundamental difference between the thresholding and optomics methods lies in the extraction of image features for tissue classification. While thresholding uses pixel intensities for tumor differentiation, optomics exploit inter-pixel spatial relationships and higher dimensional image textures to achieve the same end. Optomics provided a statistically significant improvement in classification accuracy and FNR. This improvement in prediction performance is a proof-of-concept for how spatial and textural image features enhance tissue type identification in fluorescence molecular imagery.

Threshold classification accuracy varied widely across testing set specimen images (std = 7.6%). This may be due to inter-patient or inter-specimen biases or differences in fluorescence signals between dose groups. Fluorescence image normalization reduced variations in inter-patient signals and amplified differences in normal and tumor fluorescence; however, inhomogeneous overexpression of EGFR in tumor and high endogenous expression of EGFR in normal tissues still have confounding effects on fluorescence signals, which is evident in the overlapping regions of intensity histograms from ground truth tumor and normal tissue ROIs displayed in Supplementary Figure S7. Inability to separate fluorescence signals between normal tissues and tumor caused the low classification accuracy of the thresholding method.

The optomics method was less affected by variations in imaging agent dose and inter-image bias based on the higher FPR and FNR achieved (Supplementary Figure S4B). False positive predictions by the thresholding method were mainly located in normal tissues that had high endogenous EGFR expression. This result is observed in Figure 6, where visualized predictions demonstrate that optomics successfully identified normal EGFR expressing mucosa, found along the edges of the specimens, while the fluorescence thresholding method failed to identify these regions accurately. False negative predictions by the thresholding method were found in tumor tissues with low or heterogeneous expression of EGFR (i.e., tumor necrosis in the interior of ROIs). Optomics identified these low fluorescing tumor tissues more effectively, which decreased the overall false negative predictions compared to fluorescence thresholding. Decreases in FPR and FNR demonstrate the potential value of incorporating spatial information when detecting tumor tissues.

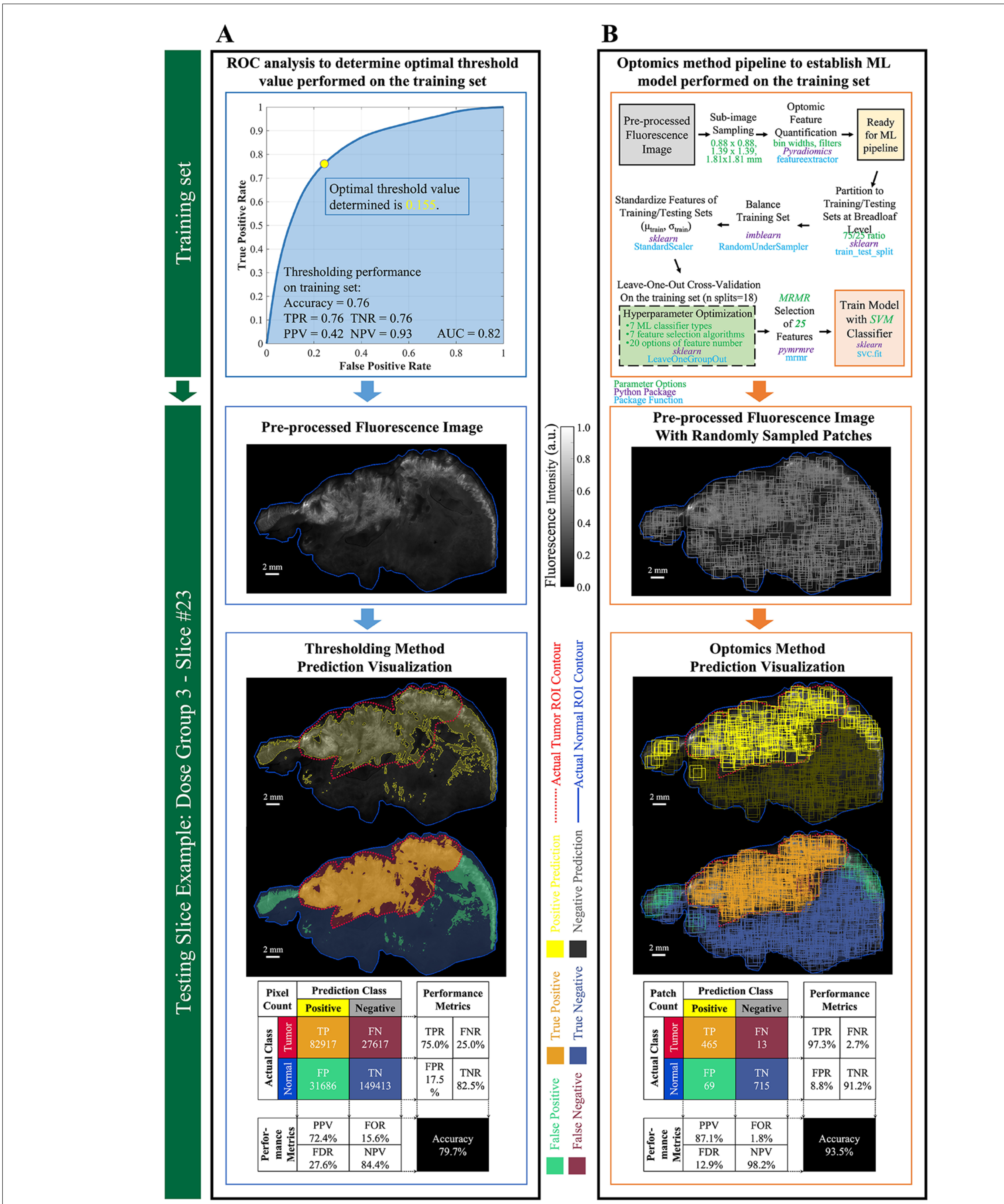

FIGURE 6
Performance evaluation of thresholding vs. optomics methods. (A) Thresholding method and its classification performance. The optimum cutoff point (i.e., threshold value) was determined through ROC curve analysis on the training set, and its predictive performance was assessed on the testing set. Pixel-level classification of the thresholding method is visualized on a testing slice example. Metrics describing its predictive performance on this testing slice example are summarized in the confusion matrix. (B) Optomics method and its classification performance. The ML model was determined based on the training set and a supervised ML pipeline summarized in the flow chart. Patch-level classification of the optomics method is visualized on the same testing slice, and its predictive performance is summarized in the confusion matrix.

Visualization of tumor extent and its boundary with normal tissues is critical in FGS and has been investigated extensively (16,

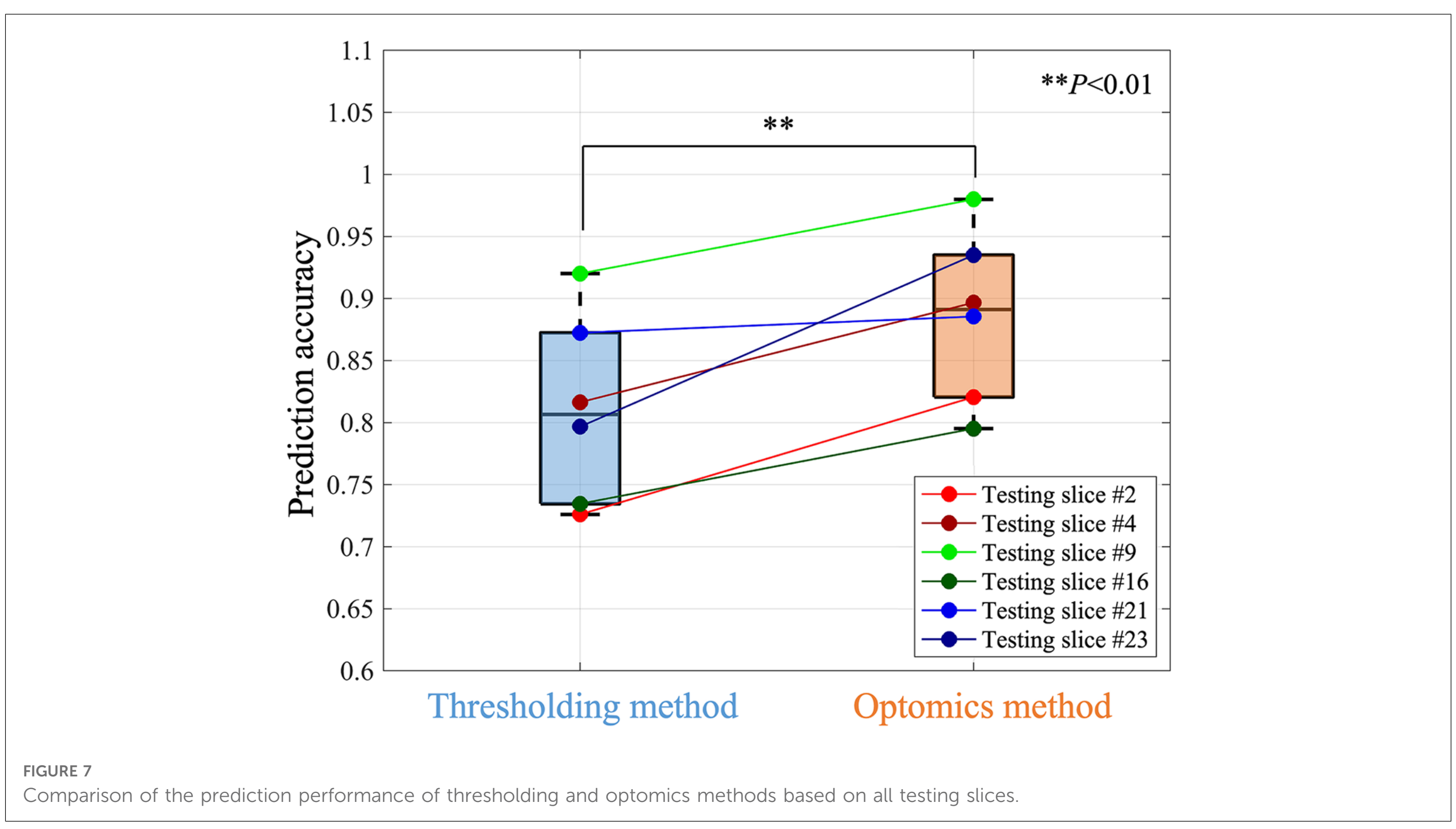

FIGURE 7
Comparison of the prediction performance of thresholding and optomics methods based on all testing slices.

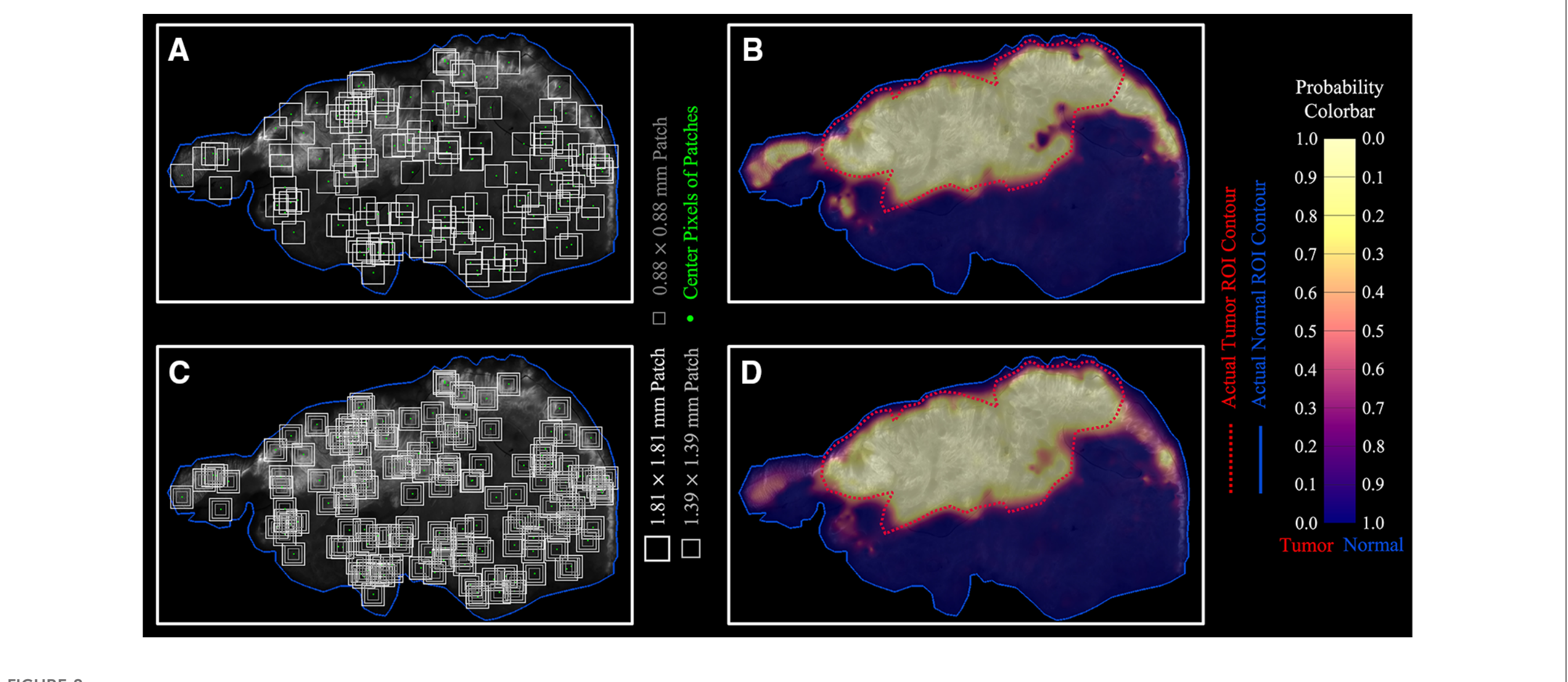

FIGURE 8
Optomics method predicted continuous probability maps. (A,B) Single patch size sampling and its probability map. (C,D) Three patch size sampling and the combined probability map based on the average probability derived from the three patch sizes.

50–53, 63). Here, a continuous tumor probability map was generated at the pixel level by interpolating predicted probabilities from center pixels of sampled patches. Production of probability maps from optomics analysis may provide a decision guide for surgeons. The probability output may be able to map the surface of the entire surgical specimen surface to identify areas suspicious for residual disease. Ideally, surgeons would threshold probability maps based on confidence intervals, their experience, and the desired goals for surgery. Probability outputs from ML models may be able to assist surgical teams during HNSCC resections and thereby help achieve improved patient outcomes.

The small size of the dataset in this study limits the generalizability of the trained optomics model as a classification tool. Furthermore, the limited number of specimens led to the use of specimen-level partitioning across training and testing sets. In the future, more data from later Phase 1 experiments would allow patient-level rather than specimen-level partitioning. Comparisons between thresholding and optomics methods also introduced bias because of disparities in the data sampling; the former performed pixel-level classification while the latter produced patch-level classification. Generation of an interpolated pixel-level probability map by averaging across optomics results from difference patch sizes may mitigate the influence of these differences, but optomics classification performance on a pixel-level was beyond the scope of this study.

This study advances an optomics analysis paradigm for classifying fluorescence image data in HNSCC specimens. Relative to conventional fluorescence intensity thresholding, optomics increased the accuracy of tumor identification and provided more precise tumor mapping for surgical guidance. Optomic signatures have potential to reveal underlying image features that are not apparent to the human observer. This preliminary assessment of optomics for the detection of malignant tissues in ABY-029 fluorescence images of HNSCC specimens suggests that extending the approach to fluorescence molecular imaging data offers value over conventional fluorescence image intensity thresholding for cancer detection during FGS.

## Data availability statement

The original contributions presented in the study are included in the article/**Supplementary Material**, further inquiries can be directed to the corresponding author/s.

## Ethics statement

The studies involving human participants were reviewed and approved by Dartmouth-Hitchcock Health Human Research Protection Program. The patients/participants provided their written informed consent to participate in this study.

## Author contributions

KSS, KDP, BWP, YC and SSS conceptualized the study. YC, SSS, BH and KSS designed the study. KSS, HSS, JRG and LJT performed the experiments. JAP performed the surgery. YC, SSS, KSS, BH and HSS analyzed the data. BWP, KDP and KSS provided supervision. SSS, BWP, KDP and KSS provided the funding. YC drafted the manuscript, and all authors reviewed the manuscript. All authors contributed to the article and approved the submitted version.

## Funding

This study was funded by the National Cancer Institute (NCI) R37 Early Investigator MERIT Award (CA212187) and F31 fellowship grant (CA257340). ABY-029 manufacturing was funded by NCI R01 CA167413. ABY-029 was developed through an academic-industrial partnership, funded through CA167413 with Affibody AB and LI-COR Biosciences, Inc. The funders were not involved in the study design, collection, analysis, interpretation of data, the writing of this article or the decision to submit it for publication.

## Acknowledgments

We would like to acknowledge Sally B. Mansur for clinical and research support.

## Conflict of interest

The authors declare that the research was conducted in the absence of any commercial or financial relationships that could be construed as a potential conflict of interest.

## Publisher's note

## Supplementary material

The Supplementary Material for this article can be found online at: https://www.frontiersin.org/articles/10.3389/fmedt.2023.1009638/full#supplementary-material.